\pdfoutput=1

\documentclass[11pt]{article}

\usepackage[]{emnlp2021}

\usepackage{times}
\usepackage{latexsym}

\usepackage[T1]{fontenc}
\usepackage[utf8]{inputenc}

\usepackage{microtype}

\usepackage{url}
\usepackage{multirow}
\usepackage{booktabs}
\usepackage{graphicx,xcolor}
\usepackage{amsmath}
\usepackage{cleveref}
\usepackage{todonotes}
\usepackage[toc,page]{appendix}

\usepackage{etoolbox}

\definecolor{hblue}{RGB}{0, 129, 105}
\definecolor{hred}{RGB}{159, 1, 98}
\definecolor{hall}{RGB}{176,196,222}

\newcommand{\squishlist}{
 \begin{list}{$\bullet$}
  { \setlength{\itemsep}{0pt}
     \setlength{\parsep}{3pt}
     \setlength{\topsep}{3pt}
     \setlength{\partopsep}{0pt}
     \setlength{\leftmargin}{1.5em}
     \setlength{\labelwidth}{1em}
     \setlength{\labelsep}{0.5em} } }

\newcommand{\squishend}{
  \end{list}  }
  
\title{Learning to Reason for Text Generation from Scientific Tables}
\author{Nafise Sadat Moosavi$^1$, Andreas R{\"u}ckl{\'e}$^1$, Dan Roth$^2$, Iryna Gurevych$^1$\\ \\ $^1$UKP Lab, Technische Universit\"at Darmstadt\\ $^2$Department of Computer and Information Science, UPenn\\  $^1$\url{https://www.ukp.tu-darmstadt.de} \\ $^2$  \url{https://www.seas.upenn.edu/\~danroth/}}

\date{}

\begin{document}
\maketitle
\begin{abstract}
    In this paper, we introduce SciGen, a new challenge dataset for the task of reasoning-aware data-to-text generation consisting of tables from scientific articles and their corresponding descriptions.  
	Describing scientific tables goes beyond the surface realization of the table content and requires reasoning over table values.
	The unique properties of SciGen are that (1) tables mostly contain numerical values, and (2) the corresponding descriptions require arithmetic reasoning.
	SciGen is therefore the first dataset that assesses the arithmetic reasoning capabilities of generation models on complex input structures, i.e., tables from scientific articles.  
	We study the effectiveness of state-of-the-art data-to-text generation models on SciGen and evaluate the results using common metrics as well as human evaluation. Our results and analyses show that (a) while humans like to reason for describing scientific tables, the ability of state-of-the-art models is severely limited on this task, (b) while adding more training data improves the results, it is not the solution for reasoning-aware text generation, and (c) one of the main bottlenecks for this task is the lack of proper automatic evaluation metrics. The data, code, and annotations for human evaluation will be available at \url{https://github.com/UKPLab/SciGen}.        
	SciGen opens new avenues for future research in reasoning-aware text generation and evaluation.

\end{abstract}

\section{Introduction}
Data-to-text generation is one of the established tasks in NLP in which the input is structured data like tables and the output is a text that describes the data. %
In this paper, we introduce a new data-to-text generation dataset, called SciGen, that contains pairs of scientific tables and their corresponding descriptions.
The descriptions of a large number of tables in the scientific articles of the computer science domain require one or more types of arithmetic reasoning over table values---e.g., argMax, argMin, comparison, subtraction, etc---over table values, indicating that humans like to reason for describing scientific tables.\footnote{For instance, we randomly select 30 tables from different articles in this domain, and 60\% of these tables were described by using arithmetic reasoning.}
Therefore, generative models that can describe scientific tables should be able to perform arithmetic reasoning.  
For instance, such models can be used to automate scientific writing by generating the result section of the article using the table of experimental results, or for developing specialized chatbots that can generate answers based on the content of tables.

In this direction, we have created SciGen to enable the development and evaluation of generation models with arithmetic reasoning capabilities.
For creating SciGen, we select tables and their corresponding descriptions from the computer science articles with the following properties: (1) the tables mostly contain numerical values, and (2) the corresponding descriptions are the result of arithmetic reasoning over table values.

We release the dataset in three settings based on the size of the training data. 
The \emph{few-shot} setting contains table-description pairs that are annotated by experts.
Since expert annotation is costly and does not scale to large data sizes, we introduce automatic pairing and pruning techniques to  extract high-quality table-description pairs in an unsupervised way.
We extend the expert annotated pairs in the few-shot setting using the automatically extracted pairs to create the medium and large splits of the dataset.
The test data is the same in all the settings and contains expert-annotated pairs.
	
We study state-of-the-art data-to-text generation models---including BART \citep{lewis-etal-2020-bart} and T5 \citep{JMLR:v21:20-074} pretrained language models---on SciGen, and we evaluate the results using common evaluation metrics as well as human evaluation.
Our results show that (1) none of the common metrics can properly discriminate the outputs of reasoning-aware text generation, and the lack of a proper automatic evaluation metric is the main bottleneck for the progress for this task, (2) while the generated outputs by the examined models are coherent and fluent and look like valid descriptions, they are mostly factually incorrect given the content of the table, i.e., the entities that appear in the description are relevant to the table but the lack arithmetic reasoning capabilities results in incorrect statements about the present results in the table, and (3) the addition of automatically extracted training data in the \emph{medium} and \emph{large} settings reduces the hallucination and improves the correctness of the generated descriptions, however, the extent that they can improve the correctness is limited, e.g., the correctness of the outputs in the \emph{large} setting is lower than those in \emph{medium}.

Overall, the main contributions of this paper are:
\begin{itemize}
	\item We release SciGen, the first reasoning-aware data-to-text generation dataset based on scientific articles. %
		\item We provide our code for extracting table-description pairs from latex files of scientific articles that (1) provides high-quality unsupervised training data, and (2) facilitates future annotation studies for new domains.
	\item We provide all the data and annotations that is used for our human evaluations to help the creation of new metrics for this task as our results suggests that existing automatic metrics are not discriminative for this task. 
	\item We show that text generation using arithmetic reasoning is a challenging task that cannot be solved by merely adding more data or applying large pre-trained models.    
\end{itemize}     

\begin{table*}[htb]
	\footnotesize
	\centering
	\begin{tabular}{lrrrrrlll}
		\footnotesize
		\textbf{Dataset} & \textbf{Pairs} & \textbf{Cell} & \textbf{Num.} & \textbf{$|$Text$|$} & \textbf{$|$Vocab$|$} & \textbf{Domain} & \textbf{Source} & \textbf{Reasoning}   \\ \midrule
		WikiBIO & 400K & 17 & 3 & 97 & 400K & Biography & Crawled & No \\
		Rotowire & 11.3K & 649 & 429 & 337 & 11.3K & Basketball & Crawled & Few \\
		ToTTo & 136K & 3 & 1 & 17 & 136K & Open (Wikipedia) & Annotated & Few \\
		LogicNLG & 37K & 91 & 35 & 14 & 122K & Open (Wikipedia) & Annotated & Yes  \\
		\textbf{SciGen} & 1.3K & 53 & 34 & 116 &  11K & Scientific & Annotated & Yes \\ 
	\end{tabular}
\caption{Comparison of SciGen to recent table-to-text generation datasets. \emph{Pairs} shows the number of annotated pairs in each dataset. The \emph{Cell} and \emph{Num.} columns show the average number of total cells and cells with numerical values in the input table, respectively. \emph{$|$Text$|$} reports the average numbers of words in descriptions. \emph{$|$Vocab$|$} is the length of the corresponding vocabulary in each dataset.    }
\label{tab:sciGen-vs-others}
\end{table*}
\section{Related Work}

The task of data-to-text generation is to generate coherent, relevant, and meaningful natural language text that describes the non-linguistic input data like tables, knowledge bases, tuples, or graphs \citep{reiter2000building,gatt2018survey}.
Existing datasets for data-to-text generation cover various domains and applications including sport reports \citep{wiseman-etal-2017-challenges,sportsett,van-der-lee-etal-2017-pass}, weather reports or forecast \citep{belz_2008,balakrishnan-etal-2019-constrained}, restaurant descriptions \citep{dusek.etal2020:csl,oraby-etal-2018-controlling,reed-etal-2018-neural}, biographies \citep{lebret-etal-2016-neural,nema-etal-2018-generating}, entity descriptions \citep{W18-6532,wang-etal-2018-describing}, as well as open-domain datasets \citep{gardent-etal-2017-webnlg,parikh2020totto}.

The textual descriptions in the majority of existing datasets mostly contain a verbalized summary of the content in the data, and is therefore a surface-level summary of the data \citep{chen2008learning,belz-etal-2011-first,lebret-etal-2016-neural,gardent-etal-2017-webnlg,dusek-etal-2018-findings,koncel-kedziorski-etal-2019-text,radev2020dart,parikh2020totto}.
SciGen, on the other hand, goes beyond the surface realization of the input data and requires arithmetic reasoning for text generation.
The most related dataset to SciGen is LogicNLG \citep{chen-etal-2020-logical}, in which the text generation step also requires logical reasoning. 

LogicNLG is based on TabFact \citep{2019TabFactA} that is a table-based fact verification datasets.
For creating TabFact, annotators were asked to write refute and entailment statements based on Wikipedia tables.
The statements were classified into simple and complex: simple statements are verifiable without involving logical inference, and complex statements involve multiple rows of the tables as well as logical operations such as summary, argMax, argMin, count, comparison, average, etc. 
LogicNLG contains the complex statements of TabFact that are labeled as entailment given their corresponding table.

Apart from their domains---i.e., Wikipedia vs.\ scientific texts, there are two main differences between LogicNLG and SciGen.
First, annotators of TabFact were asked to generate multiple statements per table. As a result, each text only describes a part of the table---i.e., on average two rows of the table---and it often only contains one type of reasoning.
The relevant rows of the table for each text are identified automatically in LogicNLG, and since identifying the relevant rows is not a trivial task, the LogicNLG examples are noisy.
SciGen, on the other hand, only contains one description per table and it may contain multiple types of reasoning. 
SciGen is therefore more challenging than LogicNLG based on both data complexity and text complexity. For instance, LogicNLG descriptions contain 14 words on average, compared to 116 words in SciGen. 

Second, the types of logical operations that are used for creating TabFact, and therefore LogicNLG, are not limited to arithmetic operations. Based on \citet{2019TabFactA}'s analyses, count is the most common logical operation in TabFact's complex statements. However, it also contains other types of reasonings like temporal reasoning---e.g., about 1200 textual descriptions in LogicNLG are generated based on before/after operations. SciGen, on the other hand, is only targeted for evaluating arithmetic reasoning.

Table~\ref{tab:sciGen-vs-others} compares SciGen with recent table-to-text generation datasets based on various properties, in particular, (a) \emph{data complexity} that is measured by the average number of containing cells in each table, (b) \emph{text complexity} that is measured by the average number of words and the size of the vocabulary in the target text, and (c) the \emph{reasoning} requirement to generate target texts.

\begin{figure*}[!htb]
\begin{minipage}{0.55\linewidth}
\includegraphics[width=\textwidth]{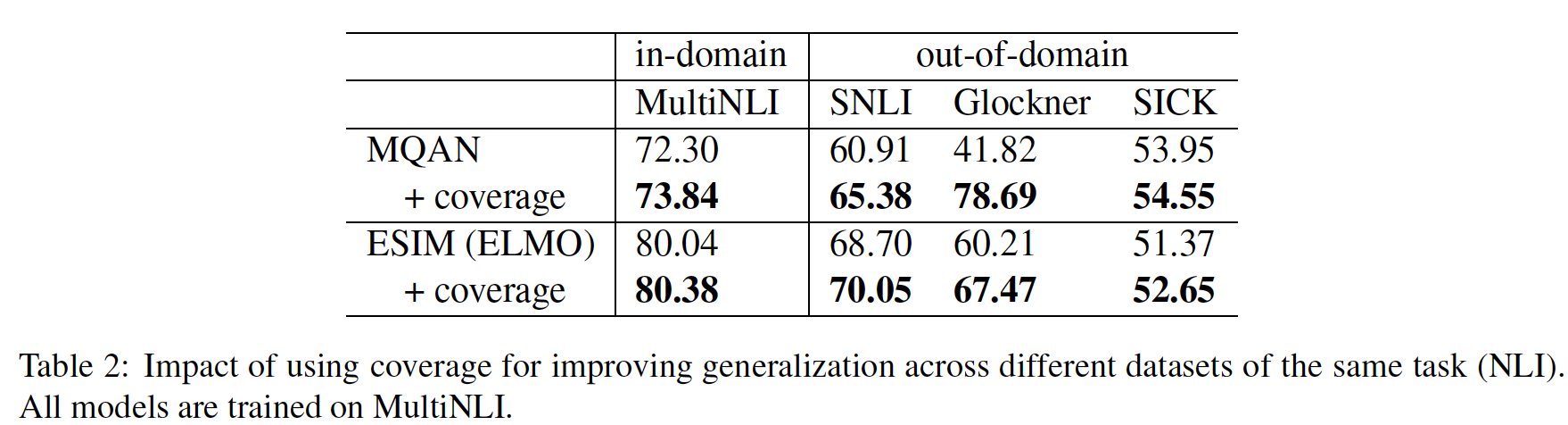}
\end{minipage}
\begin{minipage}{0.35\linewidth}
\includegraphics[width=\textwidth]{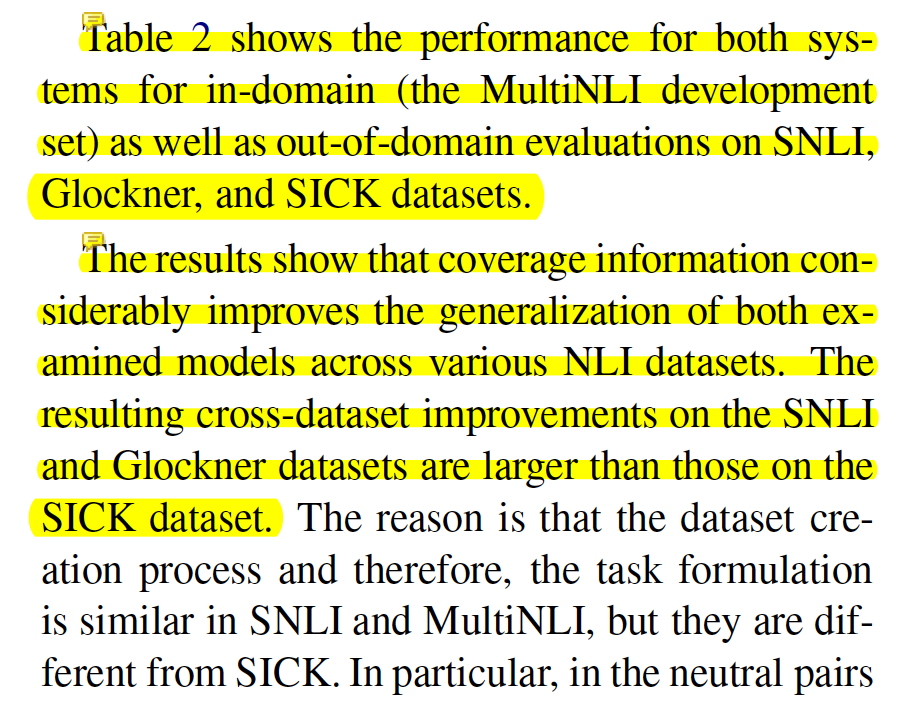}
\end{minipage}
\caption{An example of an expert table-description annotation. The annotation process contains marking the text spans from the article that (1) describes the table, and (2) can be generated using the table and its caption. }
\label{fig:man_annotation}	
\end{figure*}
\section{Dataset and Problem Definition}
\label{sect:man_anno}

\begin{table*}[htb]
	\footnotesize
\resizebox{\textwidth}{!}{%
	\begin{tabular}{lrrrrrrrr}
		\textbf{Domain} & \textbf{Article} & \textbf{Pairs} & \textbf{Cell} & \textbf{Num.} & \textbf{Cap. Sent}& \textbf{Cap. Word} & \textbf{Desc. Sent}& \textbf{Desc. Word}  \\ \midrule
		Computation and Language & 299 & 792 & 54.6 &  35.7  & 2.5  & 35.1  & 5.7  & 113.6  \\ 
		Machine Learning & 191 & 410 & 49.9 & 32.1 & 2.4 & 32.5 & 6.0 & 117.2\\
		Others & 57 & 136 & 60.1 & 36.7 & 1.9 & 25.5 & 6.1 & 126.9  \\ 

	\end{tabular}
}
\caption{The statistic of the expert-annotated data in SciGen. All domains consist of the ``Computer Science'' articles from arXiv.org. The \emph{Article} and \emph{Pairs} columns show the number of annotated articles and tables in each domain, respectively.  \emph{Cell} shows the average number of cells in the annotated tables, and \emph{Num.} shows the average number of cells containing numerical values. The \emph{Cap. sent} and \emph{Cap. word} columns show the average number of containing sentences and words in captions, respectively. Similarly, \emph{Desc. sent} and \emph{Desc. word} report the average number of sentences and words that are included in the annotated descriptions.\footnote{We use spaCy (\url{https://spacy.io/}) for computing the number of sentences and words.}  
}
\label{tab:sciGen-stat}
\end{table*}

\subsection{Problem Definition}
\label{sect:def}
SciGen is a dataset for generating descriptions from scientific tables by reasoning over their content.
An input in SciGen is a table $T$ extracted from a scientific article with its corresponding caption $C$, which is a word sequence containing one or few sentences about the table. 
$T=\{R_1, \dots,R_n\}$ is represented as a list of lists, in which each $R_i$ represents a row of the table.
The task is to generate textual description $D$ for the table that describes the most important findings of $T$ by reasoning over its content.

For instance, Figure~\ref{fig:man_annotation} shows a table from SciGen alongside its annotated description.\footnote{The table and description are from \citet{moosavi2019improving}.} As we see, generating the given description requires comparing the values of ``MQAN'' and ``ESIM (ELMO)'' rows with their corresponding ``+coverage" rows, as well as comparing the values of ``SNLI'', ``Glockner'', and ``SICK'' columns.

We can generate other descriptions from this table. For instance, ``scores of all systems are higher in SNLI compared to SICK'', or, ``except for MQAN+coverage, the rest of examined models have higher scores on SNLI compared to Glockner'', etc.
However, since the table is about evaluating the impact of coverage features, as mentioned in the caption, only a specific subset of values in the table are used for generating the description. This indicates that, apart from reasoning, relevant \emph{content selection} is also an important step in generating scientific descriptions.

\subsection{Annotation Process}
\label{man_anno}
For creating SciGen, we have selected scientific articles from arXiv.org that are accompanied by their corresponding latex sources. 
The selected articles are mainly from ``Computation and Language'' and ``Machine Learning'' fields of ``Computer Science''.
We extract tables from the selected articles using the AxCell tool \citep{axcell}.
Expert annotators---i.e., PhD students or postdocs in the corresponding field, and in many cases one of the co-authors of the selected papers---have annotated the description of each table as follows: (1) the description is selected from the text of the article, i.e., text spans of the article that describe the content of the table, and (2) only text spans that can be generated given the table and its caption are annotated as descriptions. For instance, the ``The reason is that the dataset creation process, and therefore the task formulation is similar in SNLI and MultiNLI, but are different from SICK.'' text span in Figure~\ref{fig:man_annotation} provides additional insights about the results of the table. However, it is not annotated because it cannot be generated given the table and caption.   

Two expert annotators then examine the resulting annotations from the first step and remove incorrect annotations---e.g., pairs in which the table was not extracted correctly by the tool or the annotated description was incorrect---and those in which the description does not involve reasoning, e.g., descriptions like ``The dataset statistics are summarized in Table 2''.\footnote{This step results in removing 970 table-descriptions.}

\subsection{Dataset Statistics}
Table~\ref{tab:sciGen-stat} presents the statistics of the expert-annotated table-description pairs in SciGen. The table shows the statistics for different domains of the dataset, i.e., ``Computation and Language'', ``Machine Learning'', and ``Others''.
The articles of the ``Others'' domain belong to various fields like ``Computational Geometry'', ``Distributed, Parallel, and Cluster Computing'', and ``Networking and Internet Architecture''.

\subsection{Extending the Dataset with Automatic Annotations}
Annotating descriptions of tables in scientific articles requires expert knowledge about the content of the article and is very time-consuming. 
Therefore, there is a limit to the amount of training data that can be created using expert annotations.
In this section, we propose an automatic table-description extraction process using the corresponding latex files of scientific articles. 
Using latex sources, we can easily locate paragraphs of the article that include a reference to a table.
This way, we can collect an arbitrarily large number of table-description pairs from any scientific domain.
\begin{table*}[!htb]
\centering
\footnotesize
\begin{tabular}{lrr|rrrrr}

 & \textbf{Pairs} & \textbf{$|$Text$|$} & \textbf{BLEU} & \textbf{METEOR} & \textbf{MScore}  & \textbf{BertS} & \textbf{BLEURT}\\ \hline
 automatic annotations & 950 & 182 & 31.38 & 0.64 & 0.37 & 0.90 & -0.34\\
 +post-processing  & 380 & 123 & 48.36 & 0.70 & 0.44 & 0.92 & -0.13\\
\end{tabular}
\caption{Comparing the similarity of the automatically extracted table-description pairs, before and after post-processing, to the expert annotations based on BLEU, METEOR, MoverScore, BertScore, and BLEURT metrics. \emph{Pairs} shows the number of common tables in the expert and automatically extracted annotations. \emph{$|$Text$|$} shows the average number of words in descriptions.}
\label{table:man-vs-aut-data-evaluation}
\end{table*}

We then post-process the automatically collected data by discarding pairs in which (a) the description does not reason over the table content---for this step, based on the retained and discarded pairs in the data-cleaning step of the expert annotations, we construct a set of heuristics rules for determining reasoning-aware descriptions based on captions and descriptions---, (b) the table does not contain numerical values, (c) the description describes multiple tables or figures, (d) the table is not extracted correctly by the Axcell tool, and (e) the description is too short---i.e., less than 15 words---or too long, i.e., longer than 400 words.
Additionally, we shorten the paragraphs that consist of multiple subsections to only contain the one that relates to the target table.

To assess the quality of the resulting data, we automatically extract table-description pairs from the articles in our expert-annotated data and compare the automatically extracted descriptions with those that are manually annotated.
Based on this comparison: 
\squishlist
\item 20\% of these pairs in the automatically extracted data do not exist in the expert annotations. Based on our analysis, most of these pairs were discarded from the expert annotations as a result of errors in the pdf annotation extraction tool.\footnote{Expert annotators have annotated the descriptions on the PDF files in the annotation process of \S\ref{man_anno}.}
\item 27\% of the pairs in the expert annotations do not exist in the automatic data because they were discarded by our heuristic rules in post-processing.
\item Assuming $C$ is the set of common tables in expert and automatic annotations. In the majority of the tables in $C$---i.e., 71\%---automatically extracted descriptions contain the annotated descriptions by experts. The average length of descriptions in the expert annotations of $C$'s tables is 95 words while it is 113 words for automatic annotations. This indicates that automatically extracted descriptions often contain additional information, e.g., the extracted paragraph may contain explanations about the reasons that a system performs better/worse in some settings, which cannot be generated only using the table content and its caption. 
\item In 29\% of tables in $C$, the automatic description is shorter than the expert annotation. Such cases include expert annotations that spread over multiple paragraphs among which only one or few have a reference to the table.\footnote{E.g., the annotation in the example of Figure~\ref{fig:man_annotation}.} In such cases, the automatic description is a partial description of the table. 
\squishend

Table~\ref{table:man-vs-aut-data-evaluation} reports the similarity of automatic annotations, before and after post-processing, to expert annotations according to the evaluation metrics of \S~\ref{sec:eval_met}. 
As we see, the post-processing step considerably improves the similarity of automatically extracted annotations with the expert annotations.

Our automatic table-description extraction and post-processing scripts (1) will make it possible to collect high-quality unsupervised table-description pairs from any new domain, for which we have access to latex files, and (2) will facilitate expert annotations by suggesting related paragraphs to each table as well as identifying potential reasoning-aware descriptions. This way, annotators will not need to read the whole article or section for finding the descriptions.

\subsection{Dataset Splits}
We release the dataset in three different settings: (1) \emph{few-shot}, (2) \emph{medium}, and (3) \emph{large}.

The data splits in \emph{few-shot} only contain table-description pairs from expert annotations. 
The training and development sets in this setting only contain pairs from the ``Computation and Language'' (C\&L) articles.
We split the test set into two ``C\&L'' and ``Other'' domains, in which the ``Other'' domain mainly contains examples from the ``Machine Learning'' (ML) articles. 

The training and development sets in the \emph{medium} setting contain those in \emph{few-shot} plus automatically extracted pairs from additional ``C\&L'' articles. 
Similarly, the training and development sets in the \emph{large} setting contain those in \emph{medium} in addition to automatically extracted pairs from additional ``ML'' articles.    
The test data is the same in all three settings. The ``Other'' test set can be considered as an out-of-domain evaluation set for \emph{few-shot} and \emph{medium}.
Table~\ref{tab:sciGen-setting-stat} reports the statistics of the three settings.

\begin{table}[!htb]
	\footnotesize
\resizebox{\columnwidth}{!}{%
	\begin{tabular}{lllll}
		\textbf{Setting} & \textbf{Domain} & \textbf{Train} & \textbf{Dev} & \textbf{Test} \\ \midrule
		\multirow{2}{*}{Few-shot}
		& C\&L &  200 & 100 & 492  \\ 
		& Others &  0 & 0 & 546 \\ 
		\midrule 
		\multirow{2}{*}{Medium}
		& C\&L & 200+13407 & 100+3352 & 492 \\ 
		& Other &  0 & 0 & 546 \\ \midrule
		\multirow{2}{*}{Large}
		& C\&L & 200+13407 & 100+3352 & 492 \\ 
		& Other & 26362 & 8677 & 546\\ 
	\end{tabular}
}
\caption{Number of table-description pairs in the training, development, and test sets of each setting. \emph{C\&L} consists of ``Computation and Language'' articles and \emph{Other} mostly contains ``Machine Learning'' articles.  
}
\label{tab:sciGen-setting-stat}
\end{table}

\begin{figure*}[!htb]
	\footnotesize
\begin{minipage}{0.45\linewidth}
	\centering 
\includegraphics[width=\textwidth]{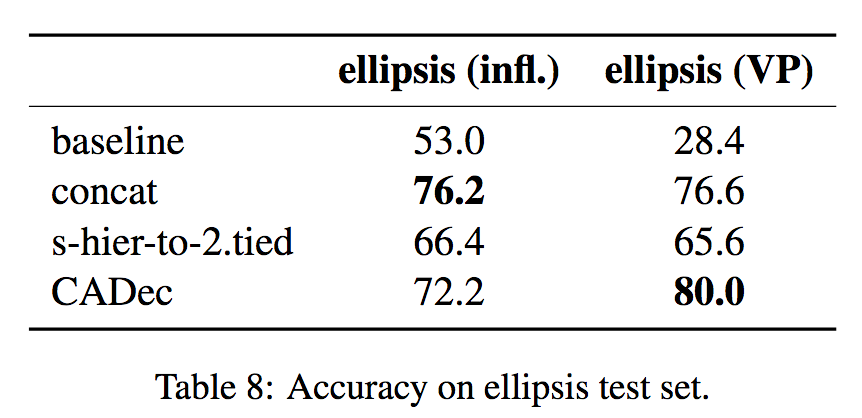}
\end{minipage}
\begin{minipage}{0.45\linewidth}
	\centering
\includegraphics[width=\textwidth]{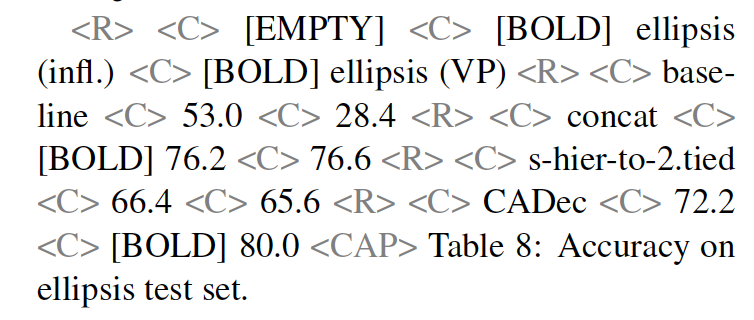}
\end{minipage}
	\caption{A sample table---from \citet{voita-etal-2019-good}---with its corresponding input representation. ``\textless R\textgreater'', ``\textless C\textgreater'', and ``\textless CAP\textgreater'' are special tokens that specify the start of the row, cell, and caption of the tables, respectively.}
	\label{fig:input_rep}
\end{figure*}
\section{Experimental Setup}
In this section, we describe the baselines, input representation format, and automatic evaluation metrics that we use in our experiments.

\subsection{Baselines}
Motivated by the results of \citet{ribeiro2020investigating} that show the BART \citep{lewis-etal-2020-bart} and T5 \citep{JMLR:v21:20-074} pretrained language models consistently outperform recent specialized data-to-text models on various benchmarks, we study the effectiveness of these two models on our dataset.

\paragraph{BART} is a Transformer-based encoder-decoder that is pretrained as a text-to-text denoising autoencoder.
For BART pre-training, input texts---taken from books and Wikipedia---are corrupted with a noising function so that the model learns to reconstruct original inputs. \citet{lewis-etal-2020-bart} show that while BART works well for comprehension tasks, it is in particular effective for text generation. 
In our experiments, we use the facebook/bart-large model from HuggingFace's Transformers \citep{Wolf2019HuggingFacesTS} with 400M parameters.

\paragraph{T5} models various NLP tasks into a unified text-to-text format, and has an on-par performance with state-of-the-art models on various NLP datasets including text classification, summarization, question answering, etc.
In our experiments, we use T5-base and T5-large models from HuggingFace's Transformers with 220M, and 770M parameters, respectively.\footnote{All models are available at \url{https://github.com/huggingface/transformers/tree/v2.10.0}}

\subsection{Input Representation}
For using text-to-text generation baselines, we should convert input tables into a text sequence.
In order to preserve the structure of the table, we use three special tokens to specify the beginning of rows, cells, and the caption of the table, namely ``\textless R\textgreater'', ``\textless C\textgreater'', and ``\textless CAP\textgreater'' tokens, respectively. Figure~\ref{fig:input_rep} shows an input table with its corresponding input representation.

\begin{table*}
	\footnotesize
\resizebox{\textwidth}{!}{%
\begin{tabular}{ll}
$D_1$ & The results of Table 2 shows that the addition of coverage features improves the performance of MQAN \\
& by 1.54, 4.47, 36.87, and 0,6 points on MultiNLI, SNLI, Glockner, and SICK, respectively. \\
&  Similarly, it improves the performance of ESIM (ELMO) by 0.34, 1.35, 7,26, and 1,28 on \\
& the mentioned datasets, respectively. We observe the highest improvements of both systems on the Glockner dataset. \\
\\
$D_2$ & The results of using coverage for MQAN and ESIM (ELMO) systems on various datasets are reported in Table 2. \\
&  The results show that the addition of coverage features significantly decrease the performance for \\
& both MQAN and ESIM (ELMO) baselines.  We observe the highest drop in the Glockner dataset. \\
\end{tabular}
}
\caption{Sample descriptions for the table in Figure~\ref{fig:man_annotation}, where $D_1$ is correct and $D_2$ includes incorrect conclusions.}
\label{sample_outputs_d1_d2}
\end{table*}

\subsection{Automatic Evaluation Metrics}
\label{sec:eval_met}
We use the following evaluation metrics:
\paragraph{BLEU} \citep{papineni-etal-2002-bleu} is one of the most common evaluation metrics for text generation. It computes the geometric average of the precision over output text's n-grams. We use SacreBLEU \citep{post-2018-call}\footnote{\url{https://github.com/mjpost/sacreBLEU}} that produces comparable and reproducible BLEU scores. 

\paragraph{METEOR} \citep{denkowski-lavie-2014-meteor} aligns the output text to the reference text and calculates sentence-level similarity scores for the alignments.

\paragraph{BertScore} \citep{bert-score} uses BERT embeddings and matches words in the output and reference sentences by cosine similarity. It then computes precision, recall, and F$_1$ measures based on the resulting matching.

\paragraph{MoverScore} \citep{zhao-etal-2019-moverscore} computes the distance between the contextual representation of the output and reference texts. It captures the amount of shared content between two texts as well as how much the output texts deviate from the reference. It uses BERT embeddings for computing contextualized representations.

\paragraph{BLEURT} \citep{sellam-etal-2020-bleurt} is a learned evaluation metric based on BERT. It is first pre-trained on synthetic examples and then fine-tuned on human judgments for the task of machine translation.

The above metrics mostly measure the surface similarity of generated descriptions to gold ones, and they cannot evaluate the factual correctness of the generated descriptions given their corresponding tables.
For instance, consider the sample descriptions in Table~\ref{sample_outputs_d1_d2}, where $D_1$ accurately describe the results while $D_2$ includes completely incorrect conclusions from the table.
The BLEU, METEOR, BertScore, MoverScore, and BLEURT scores for $D_1$ are 11.65, 0.35, 0.86, 0.27, and -0.57, respectively.
These scores for $D_2$ are 12.18, 0.30, 0.87, 0.30, and -0.54, respectively.
As we see, results for both systems based on all these evaluation metrics are in the same range, and in cases higher for the incorrect description.
As a result, we also perform human evaluation apart from reporting these standard metrics.

\section{Results}
We report the results based on both automatic evaluation metrics as well as human evaluation.
\subsection{Automatic Evaluation}
\label{sect:results}
We report the results of our baselines---i.e., BART-large, T5-base, and T5-large---on different splits of the SciGen dataset in Table~\ref{table:few-shot-medium-large} using the evaluation metrics of \S\ref{sec:eval_met}.
Based on the results: (1) except for BertScore, the value range for the rest of the metrics is very low, (2) BertScore values are very high for all the experiments, however, as we will see in \S~\ref{sect:human_eval}, generated descriptions are far from being accurate, (3) there is not a clear agreement between the rankings of different metrics for the examined models and settings, and (4) according to automatic metrics BART performs better than the two other models, however, as we will see in \S~\ref{sect:human_eval}, T5-large performs on-par with or in cases better than BART according to human evaluation.

\begin{table}[htb]
	\begin{center}
\resizebox{\columnwidth}{!}{%
\begin{tabular}{llrrrrr}
\textbf{Setting} & \textbf{Model} & \textbf{BLEU} & \textbf{METEOR} & \textbf{MoverS} & \textbf{BertS} & \textbf{BLEURT}\\ \hline
\multicolumn{7}{c}{\textbf{ Test(C\&L)}} \\ 
\hline 
\multirow{3}{*}{Few}
& BART & 4.73 & 0.22 & \textbf{0.14} & \textbf{0.84} & \textbf{-0.66} \\

&T5-base & 2.59 & 0.13 & 0.02 & 0.79 & -1.09\\
& T5-large & 3.16 & 0.16 & 0.06 & 0.81 &  -0.95\\ \hline
\multirow{3}{*}{Medium}
& BART & \textbf{5.30} & \textbf{0.23} & 0.13  & \textbf{0.84} & -0.72\\
&T5-base & 3.32 & 0.15 & 0.05 & 0.82  & -0.89\\
& T5-large & 3.65 & 0.17 & 0.10 & 0.83 & -0.77 \\ \hline 
 \multirow{3}{*}{Large}
& BART & 5.04 & 0.22 & \textbf{0.14} & \textbf{0.84} & -0.71\\
&T5-base & 3.38 & 0.15 & 0.06 & 0.82 & -0.85\\
& T5-large & 3.84 & 0.18 & 0.10 & 0.83 & -0.79 \\ 
 \hline 
 \multicolumn{7}{c}{\textbf{Test(Other) }} \\ \hline
\multirow{3}{*}{Few}
& BART &  4.26 & \textbf{0.22} & \textbf{0.13} & \textbf{0.84} & \textbf{-0.66}\\
&T5-base & 2.12 & 0.11 & 0.00 & 0.78 & -1.13\\
& T5-large & 2.58 & 0.14 & 0.03 & 0.81 & -1.01\\  \hline
\multirow{3}{*}{Medium}
& BART & {4.64} & \textbf{0.22} & 0.12 & \textbf{0.84}& -0.71 \\
&T5-base & 3.17 & 0.14 & 0.04 & 0.81 & -0.91\\
& T5-large & 3.00 & 0.16 & 0.07 & 0.82 & -0.81  \\ \hline
 \multirow{3}{*}{Large}
& BART & \textbf{4.66} & 0.20 & 0.12 & \textbf{0.84} & -0.72\\
&T5-base & 3.66 & 0.14 & 0.05 & 0.81 &  -0.86\\
& T5-large & 3.61 & 0.17 & 0.07 & 0.82 & -0.81\\ 
\end{tabular}
}
\caption{Performance of the BART-large, T5-base, and T5-large models on the \emph{few-shot}, \emph{Medium}, and \emph{Large} settings based on BLEU, METEOR, MoverScore, BertScore, and BLEURT metrics. 
We report the scores on ``C\&L'' and ``Other'' test domains separately. The ``Other'' examples are out-of-domain test data for the \emph{few-shot} and \emph{Medium} settings. Highest scores for each domain are boldfaced. }
\label{table:few-shot-medium-large}

\end{center}
\end{table}

As an example of model outputs, Table~\ref{sample_outputs} shows the outputs of the BART-large model on one of the ``C\&L'' test examples, i.e., the table in Figure~\ref{fig:input_rep}.

\begin{table*}
	\footnotesize
\resizebox{\textwidth}{!}{%
\begin{tabular}{ll}
	\toprule	
\textbf{Setting} & \textbf{Description} \\ \hline
\multirow{2}{*}{Gold} & For ellipsis, both models improve substantially over the baseline (by 19-51 percentage points), \\ 
& with concat stronger for inflection tasks and CADec stronger for VPellipsis. \\ \midrule
\multirow{2}{*}{Few} 
&  table 8 shows the performance of our model on the ellipsis test set. \\
& \textcolor{hred}{our concat model outperforms all the other models that do not use concat embeddings.} \\ \midrule
\multirow{5}{*}{Med.} 
&  \textcolor{hblue}{the results show that both concat and cadec significantly outperform the baseline and s-hier-to-2.tied} on the ellipsis test set. \\
& \textcolor{hblue}{the concat model is better than the baseline on both inflected and verb-inflated forms}, \textcolor{hred}{while} \textcolor{hblue}{cadec is better on the vp version.} \\
& \textcolor{hblue}{on the inflated version, cadec outperforms the baseline by a large margin} and \textcolor{hred}{is within 3 points of the best model.} \\
& the model with the best accuracy on the verb portion of the test set (80.0\%) \textcolor{hall}{is very close to the best performance} \\
& \textcolor{hall}{on the noun portion (75.6\%)}.  \textcolor{hred}{the results indicate that cadec has better generalizability than the other models}.\\ \midrule
\multirow{3}{*}{Large}
&  \textcolor{hall}{the concatenation model concatenates the output of the concatenated encoder and decoder,} \\
& and \textcolor{hred}{the s-hier-to-2.tied model performs slightly better than concat on both ellipsis (infl.) and vice versa}. \\
& \textcolor{hblue}{cadec outperforms the other two models by a large margin,} \\
& \textcolor{hblue}{achieving 80.0\% accuracy on the vp ellipsoids}, which is the new state of the art. \\
 \bottomrule 
\end{tabular}
}
\caption{Sample outputs from the BART model on few-shot, medium, and large settings on the corresponding table from Figure~\ref{fig:input_rep}. Factually correct statements are marked with \textcolor{hblue}{green}, factually incorrect statements and hallucinations are marked with \textcolor{hred}{red} and \textcolor{hall}{blue}, respectively. }
\label{sample_outputs}
\end{table*}

\begin{table}[!htb]
	\begin{center}
	\footnotesize
\resizebox{\columnwidth}{!}{%
\begin{tabular}{llrrrr}
Model & Setting & \textbf{Rec.} & \textbf{Prec.} & \textbf{Corr.} & \textbf{Hall.}\\ 
\hline 
\multirow{3}{*}{BART}
& Few & 0.0 & 0.0 & 0.1 & {0.4} \\
& Medium & \textbf{0.1} & \textbf{0.1} & \textbf{0.4} & 0.2\\
& Large & \textbf{0.1} & \textbf{0.1} & 0.2 & {0.3}\\ \hline
\multirow{3}{*}{T5-Large}
& Few & 0.0 & 0.0 & 0.1 & 0.3 \\ 
& Medium & \textbf{0.1} & \textbf{0.1} & \textbf{0.4} & 0.2 \\  
& Large & \textbf{0.1} & \textbf{0.1} & 0.3 & \textbf{0.1}  \\ 
\end{tabular}
}
\caption{The results of human evaluation for the BART and T5-large models according to the recall, precision, correctness, and hallucination metrics. Best values---i.e., highest coverage, precision, and correctness values and the lowest hallucination---are bold-faced. }
\label{table:human_eval}

\end{center}
\end{table}

\subsection{Human Evaluation}
\label{sect:human_eval}
For human evaluation, we select 58 table-description pairs from the SciGen ``C\&L'' test set and their corresponding system-generated descriptions from the BART and T5-large models for the three settings.\footnote{Based on our initial analysis, the quality of the descriptions of these two models were higher than those of T5-base.}
We break down each description, both gold and system-generated ones---i.e., 58$\times$2$\times$3 descriptions--to a list of individual statements. 
For instance, the corresponding statements with the gold description in Table~\ref{sample_outputs} are (a) ``For ellipsis, both models improve substantially over the baseline (by 19-51 percentage points)'', (b) ``concat is stronger for inflection tasks'', and (c) ``CADec is stronger for VPellipsis''.

We assign one of the following labels to each of the extracted statements from system-generated descriptions: (1) \emph{entailed}: a generated statement that is entailed by the corresponding gold description, i.e., is equivalent to one of the extracted statements from the gold description, (2) \emph{extra}: a statement that is not entailed by the gold description but is correct based on the table content, (3) \emph{incorrect}: a statement that is relevant to the table but is factually incorrect---e.g., ``the s-hier-to-2.tied model performs slightly better than concat on both ellipsis (infl.) and vice versa.'' in Table~\ref{sample_outputs} contains relevant entities that are mentioned in the table, but the statement is incorrect---, and (4) \emph{hallucinated}: a statement that is irrelevant to the table.

Based on the above labels, we compute four metrics as follows:

\paragraph{Recall:} the ratio of the statements in the gold description that are covered by the system-generated description, i.e., $\frac{\text{$|$entailed statements$|$}}{\text{$|$gold statements$|$}}$ per description.
\paragraph{Precision:} the ratio of the statements in the system-generated description that exist in the gold description, i.e., $\frac{\text{$|$entailed statements$|$}}{\text{$|$generated statements$|$}}$ per description.
  
\paragraph{Correctness:} the ratio of the statements in the system-generated description that are factually correct, i.e., $\frac{\text{$|$entailed statements$|$+$|$extra statements$|$}}{\text{$|$generated statements$|$}}$.

\paragraph{hallucination:} the ratio of irrelevant statements with regard to the table that is computed as $\frac{\text{$|$hallucinated statements$|$}}{\text{$|$generated statements$|$}}$.

Table~\ref{table:human_eval} presents the results of the human evaluation. 
Based on the results, we observe that:
\squishlist
\item The addition of automatically extracted pairs in the medium and large settings improves the recall, precision, and correctness of the generated descriptions and decreases their hallucination.
\item Compared to the medium setting, the generated descriptions in the large setting contain a larger number of \emph{factually-incorrect} facts, i.e., lower correctness. This could be due to the fact that the additional table-description pairs in large are from a different domain, i.e., ``ML''. 
\item There is still a large gap for solving the SciGen dataset, i.e., in the best case, only 40\% of the generated statements are correct and only 10\% of the gold statements are covered by the generated description.
\squishend
\section{Conclusions}
In this paper, we introduce SciGen that is a challenging dataset for reasoning-aware data-to-text generation.
The input data in SciGen are tables from scientific articles and generating their corresponding descriptions requires arithmetic reasoning over table values.
Annotating scientific articles requires expert knowledge and is therefore very costly and does not scale to large data sizes.
Therefore, we provide a pipeline to extract high-quality unsupervised table-description pairs from the corresponding latex files of scientific articles. 
We release SciGen in three different settings---i.e., few-shot, medium, and large---based on the size of the available training data.
The few-shot setting and the test set contain expert-annotated table-description pairs while the training data in medium and large settings contain automatically extracted table-description pairs.
We study two state-of-the-art data-to-text generation models on SciGen and evaluate the results using both common automatic evaluation metrics as well as human evaluation.
Our results show that (1) common metrics are not reliable for evaluating reasoning-aware text generation, i.e., they do not correlate with human evaluation and they also do not agree with each other, and (2) adding automatically extracted annotations improves the correctness of the generated descriptions and reduces the hallucination, however, there is still a large gap to solve the dataset.
We ensure reproducibility at all levels by releasing our data, full extraction and post-processing pipeline, and by publishing the raw responses from our human evaluation.
 
\section*{Acknowledgements}
The authors would like to thank Xia Zeng, Dennis Kuhn, Ragini Monen, and Gisela Vallejo for their great efforts in the data collection process.
We gratefully acknowledge the help of numerous members of this research community in helping us with the annotations.
This work was possible thanks to all these contributing researchers.
We would also like to thank Michael Bugert, Jan-Christoph Kile, Ji-Ung Lee, Yevgeniy Puzikov, Kevin Stowe, and Ajie Utama for their valuable feedbacks on the paper. 
This work has been funded by the German Federal Ministry of Education and Research and the Hessian Ministry of Higher Education, Research, Science and the Arts within their joint support of the National Research Center for Applied Cybersecurity ATHENE. 
\bibliographystyle{acl_natbib}
\bibliography{paper}

\appendix
\label{sec:appendix}

\section{Upper-Bound Estimation}
To estimate an upper bound for the performance of generation models on our dataset based on automatic metrics, we randomly select 50 tables from expert-annotations. We provide these tables to a new expert annotator and ask them to describe the tables in their own words without looking at the gold descriptions or the result sections of the corresponding articles.\footnote{They had access to the background sections of the article in case the information of the table itself is not enough to describe them.}
Table~\ref{table:upper-bound} shows the scores of these 50 annotated tables compared to their corresponding gold annotations.
The high-value range for all the metrics shows that automatic metrics can acknowledge the correctness of accurate descriptions, i.e., those written by an expert. However, as we see in \S~\ref{sect:results}, they cannot discriminate imperfect descriptions.

\begin{table}[htb]
	\begin{center}
\resizebox{\columnwidth}{!}{%
\begin{tabular}{rrrrr}
 \textbf{BLEU} & \textbf{METEOR} & \textbf{MoverS} & \textbf{BertS} & \textbf{BLEURT}\\ \toprule
 66.73 & 0.614 & 0.98 & 0.99 & 0.95 \\ 
\end{tabular}
}
\caption{The automatic evaluation scores for 50 table-descriptions in which an expert has written descriptions based on table contents and without looking at the gold descriptions.}
\label{table:upper-bound}
\end{center}
\end{table}

\section{Impact of Table Captions}
Table~\ref{table:few-shot-no-caption} shows the impact of captions on automatic evaluation metrics.
The \emph{caption} row shows the results when the caption is considered as the description, i.e., evaluating captions compared to gold descriptions. The \emph{BART} and \emph{T5-large} rows show the result of these two models in the few-shot setting where the captions of the tables were excluded from the input data.
\begin{table}[htb]
	\begin{center}
\resizebox{\columnwidth}{!}{%
\begin{tabular}{llrrrrr}
 Model & \textbf{BLEU} & \textbf{METEOR} & \textbf{MoverS}  & \textbf{BertS} & \textbf{BLEURT}\\ \hline
 caption & 1.68 & 0.11 & 0.12 & 0.84 & -0.98\\
BART & 2.39 & 0.17 & 0.08 & 0.83 & -0.72 \\
T5-large & 1.50 & 0.12 & 0.06 & 0.81 & -0.98  \\ 
\end{tabular}
}
\caption{Evaluating the impact of table captions. The reported results are equivalent to the reported few-shot results in Table~\ref{table:few-shot-medium-large} but with the difference that the table captions are excluded from the input. The caption rows show the values of evaluation metrics when the caption is considered as the description.}
\label{table:few-shot-no-caption}
\end{center}
\end{table}

\begin{table*}
	\footnotesize
\resizebox{\textwidth}{!}{%
\begin{tabular}{lll}
	\toprule	
\textbf{Setting} & \textbf{Model} &  \textbf{Description} \\ \hline
\multirow{2}{*}{Gold} & & For ellipsis, both models improve substantially over the baseline (by 19-51 percentage points), \\ 
& & with concat stronger for inflection tasks and CADec stronger for VPellipsis. \\ \midrule

\multirow{6}{*}{Few} 
& BART &  table 8 shows the performance of our model on the ellipsis test set. \\
& & {our concat model outperforms all the other models that do not use concat embeddings.} \\ \\
& T5-base & \textless extra\_id\_0 \textgreater c \textgreater [ bold ] ellipsis ( infl . ) c \textgreater [ s - hier - to - 2 . tied ] \\
& & c \textgreater ( s - hier - to - 2 . tied ) c \textgreater s - hier - to - 2 . tied c \textgreater ( s - hier - to - 2 . tied ) c \textgreater 80 \% accuracy on ellipsis test set . \\ \\
& T5-large & \textless extra\_id\_0 \textgreater vp ) and concatenation ( s - hier - to - 2 . tied , infl . ) , respectively , are shown in table 8 .\\
& & as expected , concatenation improves the accuracy on the infl . ellipsis test set .\\
 \midrule
\multirow{12}{*}{Medium} 
& BART &  {the results show that both concat and cadec significantly outperform the baseline and s-hier-to-2.tied} on the ellipsis test set. \\
& & {the concat model is better than the baseline on both inflected and verb-inflated forms}, {while} {cadec is better on the vp version.} \\
& & {on the inflated version, cadec outperforms the baseline by a large margin} and {is within 3 points of the best model.} \\
& & the model with the best accuracy on the verb portion of the test set (80.0\%) {is very close to the best performance} \\
& & {on the noun portion (75.6\%)}.  {the results indicate that cadec has better generalizability than the other models}.\\ \\
& T5-base & \textless extra\_id\_0 \textgreater the s - hier - to - 2 . tied model outperforms the baseline on both infl . and vp ellipsis datasets . \\
& & cadec outperforms s - hier - to - 2 . tied on both infl . and vp ellipsis datasets . \\
& & cadec achieves the best performance on both infl . and vp ellipsis datasets . \\ \\

& T5-large & can see that both concatenation and s - hier - to - 2 . tied approaches are significantly better than the baseline . \\ & & cadec outperforms s - hier - to - 2 . tied on both inflectional and verbal ellipsis \\
& & and is comparable to the baseline on verb ellipsis . \\

\midrule
\multirow{10}{*}{Large}
& BART & {the concatenation model concatenates the output of the concatenated encoder and decoder,} \\
& & and {the s-hier-to-2.tied model performs slightly better than concat on both ellipsis (infl.) and vice versa}. \\
& & {cadec outperforms the other two models by a large margin,} \\
& & {achieving 80.0\% accuracy on the vp ellipsoids}, which is the new state of the art. \\ \\

& T5-base & cadec outperforms the baseline and s - hier - to - 2 . tied by a large margin . \\
& & cadec also outperforms s - hier - to - 2 . tied by a large margin . \\ \\

& T5-large & concatenated embeddings outperform the s - hier - to - 2 . tied baseline by 6 \% on infrequent words , and by 5 \% on vp ellipsis . \\
& & the concatenated embeddings outperform the s - hier - to - 2 . tied baseline by 10 \% on infrequent words , and by 5 \% on vp ellipsis .\\
\\

 \bottomrule 
\end{tabular}
}
\caption{Sample outputs from the examined models in Table~\ref{table:few-shot-medium-large} on the corresponding table from Figure~\ref{fig:input_rep}.}
\label{sample_outputs_all}
\end{table*}

\section{Example outputs from all baselines}
Table~\ref{sample_outputs_all} provides the output of all the examined baselines for the table in Figure~\ref{fig:input_rep}.

 \end{document}